\documentclass[10pt,twocolumn,letterpaper]{article}

\usepackage[pagenumbers]{cvpr} 

\usepackage[dvipsnames]{xcolor}

\usepackage{graphicx}

\usepackage[pagebackref=true,breaklinks=true,colorlinks,bookmarks=false]{hyperref}
\usepackage{amsmath}
\usepackage{amssymb}

\usepackage{multirow}
\usepackage{multicol}
\usepackage{diagbox}
\usepackage{xfrac}

\usepackage{comment}

\usepackage[labelformat=simple,subrefformat=simple]{subcaption}
\usepackage{caption}
\usepackage{algorithm}
\usepackage{algpseudocode}


\usepackage{listings}
\def\paperID{*****} 
\def\confName{CVPR}
\def\confYear{2026}

\title{Reflective Dialogue between Teacher and Solver Agents\\
for Video Question Answering%
\thanks{This paper serves as the technical report for the
\href{https://egocross-benchmark.github.io/}{1st Cross-Domain EgoCross Challenge}
@ \href{https://egovis.github.io/cvpr26/}{EgoVis Workshop}, CVPR 2026.}}

\author{Takuya Murakawa, Toru Tamaki\\
Nagoya Institute of Technology, 
Nagoya, Japan
}

\begin{document}
\maketitle

\begin{abstract}
\noindent Various approaches have been proposed to adapt Vision-Language Models (VLMs) to specialized domains for Video Question Answering, including fine-tuning and in-context learning. However, acquiring task-specific knowledge at the inference phase from only a small labeled support set without fine-tuning remains a challenge. In this paper, we propose a method that achieves adaptation solely through inference-time context injection. Our method first constructs a \emph{Reflective Dialogue} (RD) --- a multi-turn conversation between two agents, in which \emph{Teacher} poses each support question and delivers correctness feedback, and \emph{Solver} answers and provides visual grounding explanations (or reflections) for both correct and incorrect answers. This dialogue history is then used as context at the inference phase. Experiments on the EgoCross benchmark  demonstrate that our method outperforms both a baseline zero-shot setting and a standard in-context learning approach that passes support set examples directly, achieving 3rd place in the Open-source Track of the 1st Cross-Domain EgoCross Challenge at the CVPR 2026 EgoVis Workshop, for which this paper also serves as a technical report.

\end{abstract}

\section{Introduction}

Recent advances in large Vision-Language Models (VLMs) have dramatically improved the performance of Video Question Answering (QA) \cite{bordes_arXiv2024_VLM, kim_IEEE2024_VLM_VQA}.
However, adapting them to specialized domains --- such as surgery and industry --- remains difficult due to the variations of visual appearance of the scene in each domain.
This is problematic when only a small set of QA examples (consisting of domain, video frames, question, and answer) is provided as a \emph{support set} in advance, while the rest of the examples without answers are used as a test set during inference. Generalizing to each domain from this small support set is difficult, particularly when each domain has further smaller groups of different question types. Fine-tuning is not possible for proprietary models and often for open models due to the substantial computational resources required that are frequently unavailable in practice.

Previous work tackled this problem in the following two approaches.
In-context learning (ICL)~\cite{brown_NeurIPS2020_ICL, dong_EMNLP2024_ICL_Survey} is widely used to adapt models by directly providing support-set QA examples as context; however, simply providing a list of QA pairs as in few-shot ICL is insufficient for the model to grasp domain-specific visual patterns.
Self-reflection methods~\cite{shinn_NeurIPS2023_reflexion,madaan_NeurIPS2023_self_refine,renze_FLLM2024_self_reflection} achieve feedback-driven reasoning improvement by verbally analyzing failures and accumulating reflections as context. However, these require repeated trials on the same question-answer pair and are difficult to apply directly to a few-shot setting.

To address this challenge, as a context for few-shot ICL, we propose to use a multi-turn conversation called \emph{Reflective Dialogue} (RD) between two agents (\emph{Teacher} and \emph{Solver}) of questions of a particular type in a domain (see Figure~\ref{fig:method})%
\footnote{Code is available at \url{https://github.com/tamaki-lab/EgoCross-Reflective-Dialogue}.}.
At construction phase,
the Teacher asks the Solver all questions in a given domain and question type to build a reflective dialogue consisting of the following four turns: 1) Teacher asks the question, 2) Solver answers to it, 3) Teacher gives correctness feedback and prompts Solver to reflect, 4) Solver provides visual evidence.
At inference phase, the constructed reflective dialogue is prepended as context to the test question. This guides the Solver toward more appropriate reasoning based on other questions of the same question type of the domain.

\section{Related Work}

\subsection{Video QA and Benchmarks}

Video QA is the task of answering natural language questions about video clips, and research has primarily advanced around general-purpose benchmarks such as ActivityNet-QA~\cite{yu_AAAI2019_activitynet}.
While recent large VLMs have significantly improved accuracy on these benchmarks, understanding egocentric video presents unique challenges, including rapid viewpoint changes from the camera wearer and complex interactions between the wearer and surrounding objects and people.
Ego4D~\cite{grauman_CVPR2022_ego4d} is a representative benchmark in this area, but focuses primarily on everyday activities (\eg, cooking and cleaning), leaving model generalization to specialized domains --- such as surgery, industrial assembly, extreme sports, and animal behavior ---  insufficiently explored.
EgoCross~\cite{li_AAAI2026_egocross} is designed to fill this gap as a cross-domain Video QA benchmark spanning the four specialized domains, and we use it for evaluation in this work.

Approaches to Video QA are broadly categorized into training-time fine-tuning, and inference-time adaptation, such as in-context learning (ICL)~\cite{brown_NeurIPS2020_ICL,kim_CVPR2025_videoicl}, which are discussed below.

\subsection{Few-Shot In-Context Learning}

In-Context Learning (ICL) provides few examples as context to guide a model's predictions at inference time.
Flamingo~\cite{alayrac_NeurIPS2022_flamingo} is a pioneering multimodal VLM that demonstrated few-shot Video QA through ICL alone, without task-specific fine-tuning.
VideoICL~\cite{kim_CVPR2025_videoicl} combines similarity-based example selection with confidence-based iterative inference, achieving significant accuracy gains on out-of-distribution (OOD) Video QA benchmarks.

These methods focus on how to select and utilize in-context examples before a VLM answers, but do not incorporate correctness feedback or reflection on incorrect answers of the VLM.
Our method also operates within the ICL framework but replaces a simple list of QA pairs with a multi-turn conversation that explicitly incorporates correctness feedback and verbalized visual evidence, providing richer context for inference.

\subsection{Self-Reflection and Verbal Self-Correction}

Reflexion~\cite{shinn_NeurIPS2023_reflexion} is a framework in which an LLM agent repeatedly attempts the same task, verbally analyzes failed trials, accumulates the resulting reflections as context, and leverages them in subsequent attempts.

Other self-reflection methods~\cite{madaan_NeurIPS2023_self_refine,renze_FLLM2024_self_reflection} also
rely on a dynamic retry loop over the same example, improving performance by having the model repeatedly solve the same problem.
This incurs high inference costs per test question and makes a direct application to few-shot settings difficult.

Our method shares the idea of verbal reflection but constructs multi-turn conversations in advance and uses the conversations as a static context across test questions without any per-question retry loop.

\section{Method}

\begin{figure*}
    \centering
    \includegraphics[width=1\linewidth]{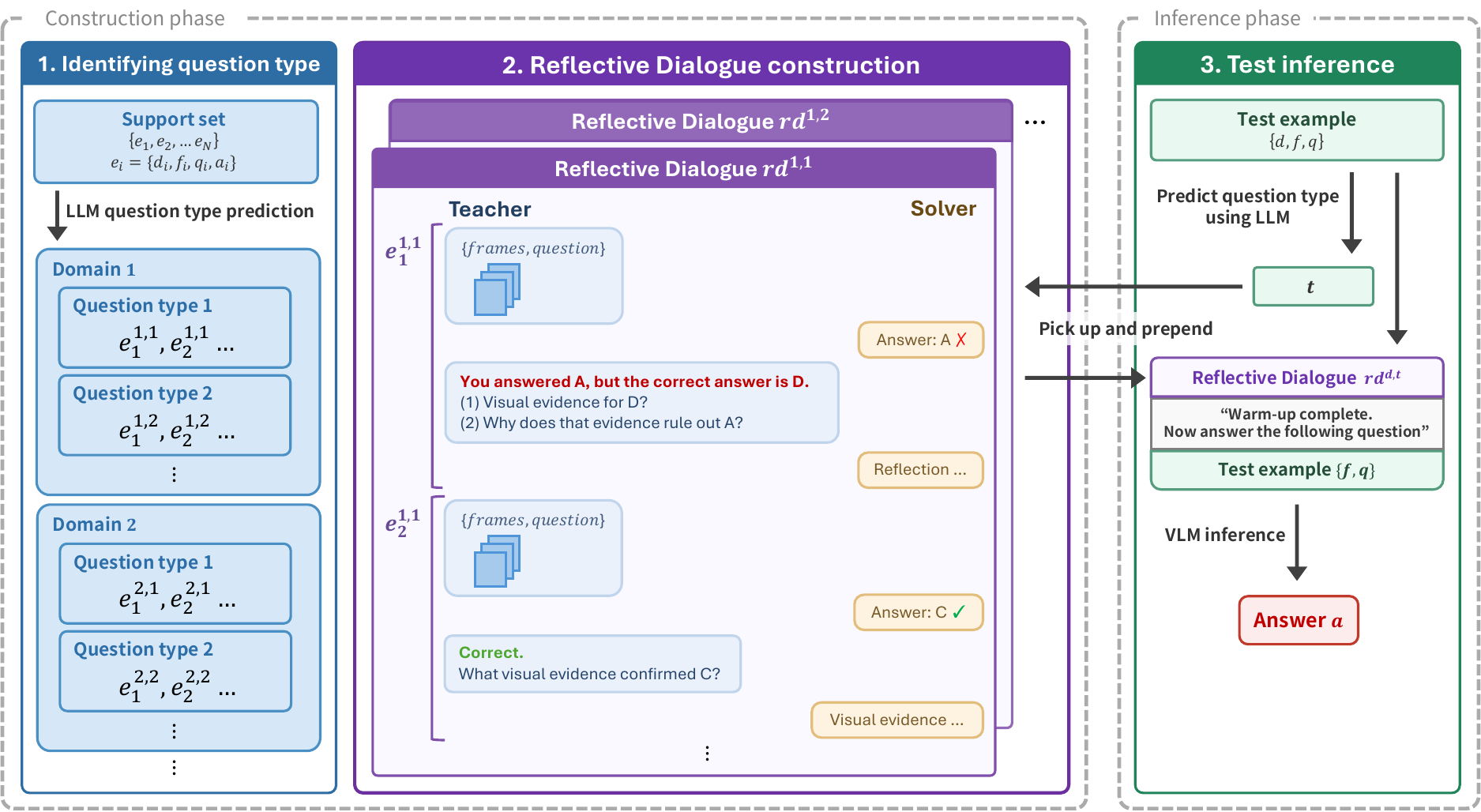}
    \caption{Overview of the proposed method.
    \textbf{(1)} Question types $t_i$ of each question $q_i$ in the support set are identified
    with LLM.
    \textbf{(2)} Reflective dialogues are constructed for each domain $d$ and question type $t$. The Solver answers the questions sequentially and the Teacher provides correctness feedback: incorrect answers trigger contrastive reflection, and correct answers prompt a visual grounding explanation. The resulting multi-turn conversation is saved as a reflective dialogue $\mathit{rd}^{d,t}$.
    \textbf{(3)} At inference phase, the corresponding reflective dialogue is retrieved and prepended to test question $q$ as a static context.}
    \label{fig:method}
\end{figure*}

The proposed method consists of the construction and inference phases (Figure~\ref{fig:method}).
In the following, we assume that the $i$-th example $e_i = \{d_i, f_i, q_i, a_i\}$ consists of domain $d_i \in \mathcal{D}=\{\mathcal{D}_1, \ldots, \mathcal{D}_D\}$,
video frames $f_i$, question text $q_i$, and answer choice $a_i \in \mathcal{A}=\{\mathcal{A}_1, \ldots, \mathcal{A}_A\}$ for 
a multi-choice question. All examples in a support set are used in the 
construction phase, while each example in a test set does not have any answer
and is used only in the inference phase.
For the EgoCross dataset \cite{li_AAAI2026_egocross}, there are $D=4$ domains and
$A=4$-way multi-choice answers are $\mathcal{A}=\{\text{``A'', ``B'', ``C'', ``D''}\}$.

\subsection{Identifying question type}
\label{sec:Identifying question type}

There are various questions in VQA datasets, such as asking object categories, relations between objects, or timing when events take place.
If such a question type $t_i \in \mathcal{T}=\{ \mathcal{T}_1, \ldots,  \mathcal{T}_T\}$ for each question $q_i$ is provided,
we use it in the reflective dialogue construction described below.

Otherwise, we use an LLM to predict question types.
We input the question text $q_i$ into the LLM and prompt it to select the most appropriate type as $t_i$.
For the EgoCross dataset \cite{li_AAAI2026_egocross}
types are one of $T=15$ subtasks
(\eg, ``action temporal localization'' or ``object spatial localization'').

In the following, we denote $e_i^{d_i, t_i}$ as an example
of domain $d_i$ with question of type $t_i$.

\subsection{Constructing reflective dialogues}

We then construct a multi-turn conversation (called reflective dialogue) between Teacher and Solver agents to solve questions and reflect results.
A reflective dialogue consists of conversation between the two agents: Teacher presents a question and provides correctness feedback, and Solver provides answers and visual evidence in cases of success or verbal reflections for failure cases.
This conversation encompasses correctness feedback and verbal descriptions on both correct and incorrect answers, which has a rich context of a particular type and domain of questions. Hence it is useful when subsequently used as context at the inference phase.
The procedure to construct reflective dialogue $\mathit{rd}^{d,t}$
of domain $d$ with question of type $t$, is the following.

For each question $e_i^{d_i, t_i}$, perform a conversation between two agents in the following four turns:
\begin{enumerate}
  \item Question turn: the Teacher provides video frames $f_i$ and the question text $q_i$.
  \item Answer turn: the Solver performs inference and outputs an answer $\hat{a}_i$.
  \item Feedback turn:
    \begin{itemize}
      \item If correct ($a_i = \hat{a}_i$): the Teacher notifies the Solver that the answer was correct and prompts it to explain in a few sentences which visual evidence confirmed the answer decision.
      \item If incorrect  ($a_i \neq \hat{a}_i$): the Teacher tells the Solver that the answer was incorrect and shows the correct answer, and prompts it to explain in a few sentences (1) which visual evidence supports the correct answer, and (2) why the incorrect answer was selected.
    \end{itemize}
  \item Reflection turn: the Solver describes the visual evidence to the answer regardless of correctness.
\end{enumerate}

By repeating the procedure for all questions $e_i^{d, t}$ sequentially (to carry over context), obtained four-turn conversations are concatenated into a single continuous conversation (a reflective dialogue $\mathit{rd}^{d,t}$), where each question's exchange immediately follows the previous one.
This dialogue includes visual evidence for both correct and incorrect answers and contrastive analyses for mistakes.
An example of a reflective dialogue is shown in Listing~\ref{list:warmup_example}.

\begin{lstlisting}[float=t,
caption={An example of reflective dialogues ($d=\text{industry}$, $t=\text{``action temporal localization''}$, two questions).
``\texttt{<frame\textit{i}>}'' is a placeholder of $i$-th video frame.},label={list:warmup_example}]
[Teacher]
<frame1><frame2>...<frameN> At what approximate timestamp did the operator's right hand first perform 'contact' with the screwdriver within the video segment from 0.00s to 15.00s?
A: 1.50s  B: 12.17s  C: 3.25s  D: 8.60s

[Solver]
The operator's right hand first performs 'contact' with the screwdriver at approximately 1.50s, as shown in the video frames.
Final Answer: A

[Teacher]
You answered A, but the correct answer is D. (1) What specific visual evidence in the frames supports D? (2) Why does that evidence rule out A? Answer in 2-3 sentences.

[Solver]
At about 8.6s the frames show the fingers closing onto the screwdriver's handle for the first time -- that is the first true 'contact', which supports D. At ~1.5s (option A) the hand only enters the frame and hovers near the tool without grasping it, so it is approach rather than contact -- this rules out A. I should distinguish hovering from the moment the fingers physically secure the tool. [...]

[Teacher]
<frame1><frame2>...<frameN> At what approximate timestamp did the operator first 'release' the screwdriver within the video segment from 0.00s to 30.00s?
A: 5.30s  B: 18.60s  C: 22.40s  D: 28.15s

[Solver]
Based on the frames, the operator first releases the screwdriver at approximately 22.40s, when the fingers open and the hand withdraws from the tool.
Final Answer: C

[Teacher]
Correct. In 1-2 sentences, what key visual evidence from the frames confirmed that C is the right answer?

[Solver]
At around 22.40s, the frames clearly show the operator's fingers spreading apart and the hand pulling away from the screwdriver, leaving it resting on the workbench. The earlier frames (5--22s) consistently showed the hand gripping the tool, making the release moment distinctly identifiable. [...]
\end{lstlisting}

\subsection{Inference with reflective dialogues}
\label{sec:Inference with reflective dialogues}

At the inference phase, the question type $t$ of a test question $q$ of domain $d$ is first identified as in Section~\ref{sec:Identifying question type}.
Then reflective dialogue $\mathit{rd}^{d,t}$ is \emph{prepended} to the test question as context, forming $[\mathit{rd}^{d,t}, q]$.
Additionally, 
a separator sentence $s=$ ``Warm-up complete. Now answer the following question'' is inserted between the reflective dialogue and the test question, $[\mathit{rd}^{d,t}, s, q]$, if necessary (depends on LLM implementations).
This inference procedure is performed for each test question independently,
preventing a pollution of subsequent inference when proprietary LLM models are used.

\subsection{Discussion}

The proposed reflective dialogue for ICL is distinct from the existing approaches.
While a standard ICL provides QA pairs as context, our method is inspired by self-reflection methods and incorporates correctness feedback and verbal reflection into the context.
Unlike self-reflection methods that require a retry loop for each test question, our method pre-constructs reflective dialogues from the support set and reuses them as a static context, avoiding per-question inference overhead.

\section{Experiments}

\subsection{Experimental setup and dataset}

We validate the proposed method in the 1st Cross-Domain EgoCross Challenge at the CVPR 2026 EgoVis Workshop.
This challenge uses the EgoCross dataset \cite{li_AAAI2026_egocross} and focuses on Video QA for egocentric videos across various specialized domains, requiring models to demonstrate generalization capabilities using a small support set.

EgoCross \cite{li_AAAI2026_egocross} is a benchmark designed to evaluate how well MLLMs can generalize to egocentric Video QA.
Unlike traditional Video QA datasets of daily-life videos, it focuses on four specialized domains (surgery, industry, xsports (extreme sports), and animal) assessing model robustness under diverse visual and semantic conditions.
It comprises 15 subtasks spanning four task categories (identification, localization, prediction, and counting), containing a total of 798 video clips and 957 question-answer pairs.
The questions are multiple-choice, each providing four possible answers.
A support set and a test set are provided for each domain.
The support set contains 20 questions per domain (80 questions in total) distributed across the 15 subtasks, with ground-truth answers provided.
The test set contains 957 questions distributed across the four domains: surgery (283), industry (245), xsports (246), and animal (183).
Performance is evaluated using CloseQA accuracy \cite{li_AAAI2026_egocross}, which measures the percentage of correctly answered multiple-choice questions.

We compare the following three methods:
\begin{itemize}
  \item \textbf{Zero-Shot}: a zero-shot setting that does not use the support set, providing only a domain-specific system prompt, and the test question with video frames.
  \item \textbf{ICL (In-Context Learning)}: a few-shot setting that prepends the support set questions, video frames, and ground-truth answers as context, without any correctness feedback or reflection.
  \item \textbf{RD (Reflective Dialogue)}: our method that prepends reflective dialogues to the test question.
\end{itemize}

\subsection{Prompt}

\noindent\textbf{System and user prompts.}
A system prompt is used to provide an expert perspective tailored to each domain (\eg, ``You are an expert analyzing egocentric video frames from surgical procedures'' for the surgery domain), and prompts the model to answer as ``Final Answer: X'' (where X is A, B, C, or D).
The reflective dialogue, separator, and test question are concatenated as described in Section~\ref{sec:Inference with reflective dialogues}
and provided as the user prompt, as the models described in Section~\ref{sec:Models} support separate system and user prompts.

\noindent\textbf{Timestamp.}
For temporal question types, we prepend a timestamp string before each frame to assist the model's temporal reasoning.
In the EgoCross dataset, video frames are provided with frame IDs in filenames, but absolute timestamps are not included.
For temporal question types that ask ``at what second did an event occur'', it is difficult for the model to answer unless a mapping between frames and timestamps is provided.
Based on the specifications of EgoCross \cite{li_AAAI2026_egocross}, we calculated timestamps
and prepended a timestamp string before each frame (\eg, \texttt{[Frame at 1.0s]<frame1>[Frame at 2.0s]<frame2>...}).

\noindent\textbf{Reduction of tokens.}
Since a reflective dialogue is prepended to test questions,
a large number of tokens directly leads to a significant increase in API costs.
For proprietary models, we limit the number of video frames $f_i$ up to 5 by uniform sampling.

\subsection{Models}
\label{sec:Models}

\noindent\textbf{Open-weight model.}
We used Qwen3-VL-4B-Instruct \cite{bai_arXiv2025_Qwen3VL}, an open-weight model.
The model is obtained from the HuggingFace Hub \cite{wolf_ACL2020_transformers}, and inference is performed on a single GPU using an NVIDIA RTX PRO 6000 Blackwell Max-Q Workstation Edition.
The input image resolution is restricted to a maximum of 128,000 pixels per frame.

\noindent\textbf{Proprietary model.}
We also used proprietary models for comparison.
We used Gemini 3.1 Pro Preview \cite{google_gemini31pro} and Gemini 3.1 Flash Image Preview \cite{google_gemini31flash}
via Google Cloud's Vertex AI, which currently achieve state-of-the-art performance on the MMMU Pro benchmark \cite{yue_ACL2025_MMMU_PRO} while maintaining lower costs compared to other commercial models.

Gemini 3.1 Pro Preview supports \emph{context caching}, which allows us to cache reflective dialogues shared across test questions, thereby reducing the cost for redundant input tokens
(we do not use caching for Gemini 3.1 Flash Image Preview as it does not support caching).

\subsection{Results}

\subsubsection{Open-weight model}

Table~\ref{tab:results_qwen} shows the per-domain and overall accuracy.
Our proposed method (``RD'' in Table~\ref{tab:results_qwen}) outperformed the baseline (``zero-shot'') in the overall metric as well as across all individual domains, demonstrating that adaptation to each domain (and question type) with the reflective dialogues is an effective context for inference-time reasoning.

``ICL'' serves as a comparative baseline to isolate the contributions of feedback and reflection; its results and related discussions are detailed in Section~\ref{subsec:ablation} of ablation study.

\begin{table*}[t]
  \centering
  \caption{Per-domain and overall accuracy of an open-weight model. Underline indicates the best among the three methods.  RD stands for our reflective dialogue.}
  \label{tab:results_qwen}
  \resizebox{\textwidth}{!}{%
  \begin{tabular}{llccccc|ccc}
    \toprule
    Model & Method & Animal & XSports & Industry & Surgery & Overall & Input Tokens (M) & Peak VRAM (GB) & Time (min) \\
    \midrule
    \multirow{3}{*}{Qwen3-VL-4B-Instruct}
      & zero-shot             & 0.541          & 0.386          & 0.331          & 0.463          & 0.424 &  1.4 & 11.23 & 15.4 \\
      & ICL & \underline{0.618} & 0.411       & 0.441          & 0.445          & 0.468 &  7.4 & 13.30 & 51.1 \\
      & RD           & 0.590          & \underline{0.419} & \underline{0.457} & \underline{0.512} & \underline{0.489} &  9.3 & 13.78 & 60.0 \\
    \bottomrule
  \end{tabular}%
  }
\end{table*}

\subsubsection{Proprietary models}

Table~\ref{tab:results_gemini} summarizes the per-domain and overall accuracy, token usage, and costs.
Across all models, Gemini 3.1 Pro Preview with RD and timestamp achieved the highest overall accuracy, setting the top score in all domains as well as the overall metric.
The addition of timestamp information led to a significant improvement in the animal, xsports, and industry domains, whereas no change was observed in the surgery domain, indicating that the effectiveness of timestamp information depends on the domain.

Notably, Gemini 3.1 Flash Image Preview with RD outperformed, in terms of the overall accuracy, the higher-tier Gemini 3.1 Pro Preview in the zero-shot setting, while incurring approximately 25\% lower costs.
This demonstrates that inference-time adaptation via reflective dialogues is effective enough to compensate for the inherent performance gap from a naive zero-shot setting.

Furthermore, for Gemini 3.1 Pro Preview with reflective dialogues, 
context caching reduced costs by approximately 49\% compared to without caching.
\begin{table*}[t]
  \centering
  \caption{Per-domain and overall accuracy, token usage, and cost of proprietary models. Underline indicates the best method within each model; bold indicates the best across all models. RD stands for our reflective dialogue.}
  \label{tab:results_gemini}
  \resizebox{\textwidth}{!}{%
  \begin{tabular}{llccccc|ccc}
    \toprule
    Model & Method & Animal & XSports & Industry & Surgery & Overall & Input Tokens (M) & Cached Tokens (M) & Total Cost (\$) \\
    \midrule
    \multirow{3}{*}{Gemini 3.1 Pro Preview}
      & zero-shot                & 0.678 & 0.455 & 0.469          & 0.668          & 0.564          & 11.9 & ---  & 23.92 \\
      & RD              & 0.716 & 0.504 & 0.535 & \underline{\textbf{0.749}} & 0.625          & 37.7 & 25.2 & 38.27 \\
      & RD + timestamps & \underline{\textbf{0.792}} & \underline{\textbf{0.524}} & \underline{\textbf{0.592}} & \underline{\textbf{0.749}} & \underline{\textbf{0.659}} & 37.7 & 25.1 & 38.30 \\
    \midrule
    \multirow{2}{*}{\parbox{3cm}{Gemini 3.1 Flash Image Preview}}
      & zero-shot    & 0.667          & 0.455          & 0.400          & \underline{0.682} & 0.549          & 12.2 & ---  &  6.09 \\
      & RD  & \underline{0.710} & \underline{0.472} & \underline{0.482} & 0.636 & \underline{0.568} & 36.1 & ---  & 18.07 \\
    \bottomrule
  \end{tabular}%
  }
\end{table*}

\subsection{Ablation study}
\label{subsec:ablation}

Here we focus on the comparison of the proposed RD with ICL, which lacks reflection.

In Table~\ref{tab:results_qwen},
both ICL and RD significantly outperformed Zero-shot, confirming the effectiveness of providing the support set as a context.
Furthermore, RD surpassed ICL in the overall metric and achieved the highest scores in three domains: xsports, industry, and surgery.
The improvement was particularly pronounced in the surgery domain. This suggests that in domains with complex procedures where extracting patterns from frames is difficult, verbalizing ``what was missed'' through reflection on incorrect answers has a stronger reinforcing effect on subsequent inference.

Conversely, in the animal domain, ICL outperformed RD and also zero-shot.
This can be interpreted as visual appearance serving as direct cues for animal identification and action recognition, thereby limiting the relative benefit of verbal reflections.

These results imply that reflective dialogues yield benefits beyond a simple presentation of few-shot question-answer pairs, although the extent of the benefit depends on the visual complexity and ambiguity of the domain.

\subsection{Comparison with fine-tuning}
\label{subsec:finetune}

Although our proposed reflective dialogue is training-free,
here we present and compare results of fine-tuning (FT) the open-weight model under several conditions: 6-epoch FT on the target domain's support set, combining FT with RD, and additional timestamp information. 
The FT configuration is shown in Table~\ref{tab:ft_config}.
We trained a single LoRA adapter \cite{hu_ICLR2021_lora} on all support set samples of the target domain for 6 epochs, without splitting by domain or question type. We utilized LLaMA-Factory \cite{zheng_ACL2024_llamafactory} for the implementation.

\begin{table}[t]
  \centering
  \caption{Fine-tuning configuration.}
  \label{tab:ft_config}

  \resizebox{0.8\linewidth}{!}{%
  \begin{tabular}{ll}
    \toprule
    Setting & Value \\
    \midrule
    Base model & Qwen3-VL-4B-Instruct \\
    Method & LoRA (SFT) \\
    LoRA rank / $\alpha$ / dropout & 64 / 128 / 0.05 \\
    LoRA target & all linear layers \\
    Optimizer & AdamW \\
    Learning rate & $1\times10^{-5}$ \\
    LR scheduler & cosine (no warmup) \\
    Epochs & 6 \\
    Batch size & 1 (per-device $1\times$ grad.\ accum.\ $1$) \\
    Precision & bfloat16 \\
    \bottomrule
  \end{tabular}
  }
\end{table}

These results are shown in Table~\ref{tab:results_qwen_ft}.
FT significantly outperformed zero-shot, with particularly notable improvements in the industry domain, where zero-shot struggles.
Combining FT with RD further improved the overall accuracy and most per-domain scores, indicating that the proposed RD acts complementarily with FT.
Even without training, RD alone approached the same performance as FT, demonstrating that much of the performance gap can be bridged solely by RD during inference without FT.
Note that in the surgery domain, the training-free RD outperformed FT, suggesting that the benefits of FT are domain-dependent.

\begin{table}[t]
  \centering
  \caption{Per-domain and overall accuracy under fine-tuning (FT) and the reflective dialogues (RD).  Underline indicates the better method within each comparison block; bold indicates the best per column across all settings.}
  \label{tab:results_qwen_ft}
  \resizebox{\columnwidth}{!}{%
  \begin{tabular}{lccccc}
    \toprule
    Method & Animal & XSports & Industry & Surgery & Overall \\
    \midrule
    zero-shot                          & 0.541 & 0.386 & 0.331 & 0.463 & 0.424 \\
    RD   & \underline{0.590} & \underline{0.419} & \underline{0.457} & \underline{\textbf{0.512}} & \underline{0.489} \\
    \midrule
    FT                                & 0.612 & \underline{\textbf{0.435}} & 0.555 & 0.484 & 0.514 \\
    FT + RD                  & \underline{\textbf{0.628}} & 0.427 & \underline{\textbf{0.588}} & \underline{0.498} & \underline{\textbf{0.528}} \\
    \bottomrule
  \end{tabular}%
  }
\end{table}

\section{Conclusion}

In this paper, we proposed Reflective Dialogue (RD), a method to adapt agents to specialized domains (and question types) by constructing a conversation between Teacher and Solver agents, which is injected into inference as context.
Our method introduces a novel approach that combines the ICL framework with offline verbal reflection, transferring knowledge across different questions in the support set rather than repeating trials on the same question.

A limitation is a substantial increase in input tokens due to prepending the reflective dialogues as a context.
Context caching can mitigate this issue, however, the number of input tokens to process remains large, inevitably increasing inference time
and API costs. A possible future work is to explore efficient context compression or selection of representative questions.

\section*{\uppercase{Acknowledgments}}
This work was supported in part by JSPS KAKENHI Grant Number JP25K03138.

{
\small
\bibliographystyle{ieeenat_fullname}
\bibliography{mybib,all}
}

\end{document}